\documentclass{article}

\usepackage{microtype}
\usepackage{graphicx}
\usepackage{subfigure}
\usepackage{booktabs} 
\usepackage{xspace}
\usepackage{makecell}
\usepackage{multirow}
\usepackage{enumitem}
\usepackage{multirow}
\usepackage{subcaption}
\usepackage[normalem]{ulem}
\useunder{\uline}{\ul}{}
\usepackage{hyperref}

\usepackage[accepted]{icml2024}

\usepackage{amsmath}
\usepackage{amssymb}
\usepackage{mathtools}
\usepackage{amsthm}

\usepackage[capitalize,noabbrev]{cleveref}

\theoremstyle{plain}

\theoremstyle{definition}

\theoremstyle{remark}

\newcommand{\method}{MPA\xspace}
\newcommand{\llms}{LLMs\xspace}
\newcommand{\gptfour}{GPT-4-Turbo\xspace}
\newcommand{\gptthree}{GPT-3.5-Turbo\xspace}
\newcommand{\gemini}{Gemini-Pro\xspace}
\newcommand{\llama}{Llama2-70b-chat\xspace}
\newcommand{\yi}{Yi-34b-chat\xspace}
\newcommand{\mixtral}{Mixtral-8x7b-Instruct\xspace}
\newcommand{\mmlu}{MMLU\xspace}
\newcommand{\arc}{ARC-C\xspace}
\newcommand{\gsm}{GSM8K\xspace}
\newcommand{\bbh}{BBH\xspace}

\newcommand{\prompt}[1]{{\small \ttfamily #1}\xspace}

\usepackage[textsize=tiny]{todonotes}

\icmltitlerunning{Dynamic Evaluation of Large Language Models by Meta Probing Agents}

\begin{document}

\twocolumn[
\icmltitle{Dynamic Evaluation of Large Language Models by Meta Probing Agents}




\begin{icmlauthorlist}
\icmlauthor{Kaijie Zhu}{ms}
\icmlauthor{Jindong Wang}{ms}
\icmlauthor{Qinlin Zhao}{ustc}
\icmlauthor{Ruochen Xu}{ms}
\icmlauthor{Xing Xie}{ms}
\end{icmlauthorlist}

\icmlaffiliation{ms}{Microsoft Research}
\icmlaffiliation{ustc}{University of Science and Technology of China}
\icmlcorrespondingauthor{Kaijie Zhu}{kaijiezhu11@gmail.com}
\icmlcorrespondingauthor{Jindong Wang}{Jindong.Wang@microsoft.com}

\icmlkeywords{LLMs, Evaluation}

\vskip 0.3in
]



\printAffiliationsAndNotice{} 

\begin{abstract}

Evaluation of large language models (LLMs) has raised great concerns in the community due to the issue of data contamination. Existing work designed evaluation protocols using well-defined algorithms for specific tasks, which cannot be easily extended to diverse scenarios. Moreover, current evaluation benchmarks can only provide the overall benchmark results and cannot support a fine-grained and multifaceted analysis of \llms' abilities. In this paper, we propose \textit{meta probing agents} (\method), a general dynamic evaluation protocol inspired by psychometrics to evaluate \llms. \method designs the probing and judging agents to automatically transform an original evaluation problem into a new one following psychometric theory on three basic cognitive abilities: language understanding, problem solving, and domain knowledge. These basic abilities are also dynamically configurable, allowing multifaceted analysis. We conducted extensive evaluations using \method and found that most \llms achieve poorer performance, indicating room for improvement. Our multifaceted analysis demonstrated the strong correlation between the basic abilities and an implicit Matthew effect on model size, i.e., larger models possess stronger correlations of the abilities. \method can also be used as a data augmentation approach to enhance \llms. Code is available at: \url{https://github.com/microsoft/promptbench}.

\end{abstract}
\section{Introduction}

Intelligence evaluation has never been as important as today due to the contradiction between unprecedented performance and underexplored interpretability of large language models (\llms)~\citep{gpt4,team2023gemini}.
To facilitate a better understanding of the strength and weakness of \llms, evaluation was carried out by collecting data from various domains~\citep{liang2022holistic,srivastava2023beyond}, benchmarking specific tasks~\citep{mmlu,humaneval,cobbe2021training,clark2018think}, and evaluating in extreme scenarios~\citep{zhu2023promptbench,wang2023decodingtrust,yang2022glue}.

There have been increasing concerns about the genuine abilities of \llms in public benchmarks, attributing the ``false promise'' to data contamination \citep{lovin2023gpt4, stochasticparrot, kocon2023chatgpt}, overfitting benchmarks~\citep{zhu2023dyval}, improper choice of the evaluation criterion~\citep{schaeffer2023emergent}, or lack of causality~\citep{berglund2023reversal}.
Among these concerns, data contamination remains the most significant, as static public benchmarks could easily be harnessed to train models.
Moreover, evaluation should provide not only benchmark results, but also insight into the structural capabilities of models for future development~\citep{burnell2023revealing}.
For example, evaluations are typically done in a certain context, e.g., a math application problem requires at least two abilities: language understanding (to comprehend the problem) and reasoning (to solve the problem).
Which ability matters more, and how can we quantify the relationship between these abilities?

\begin{figure}[t!]
    \centering
    \includegraphics[width=0.48\textwidth]{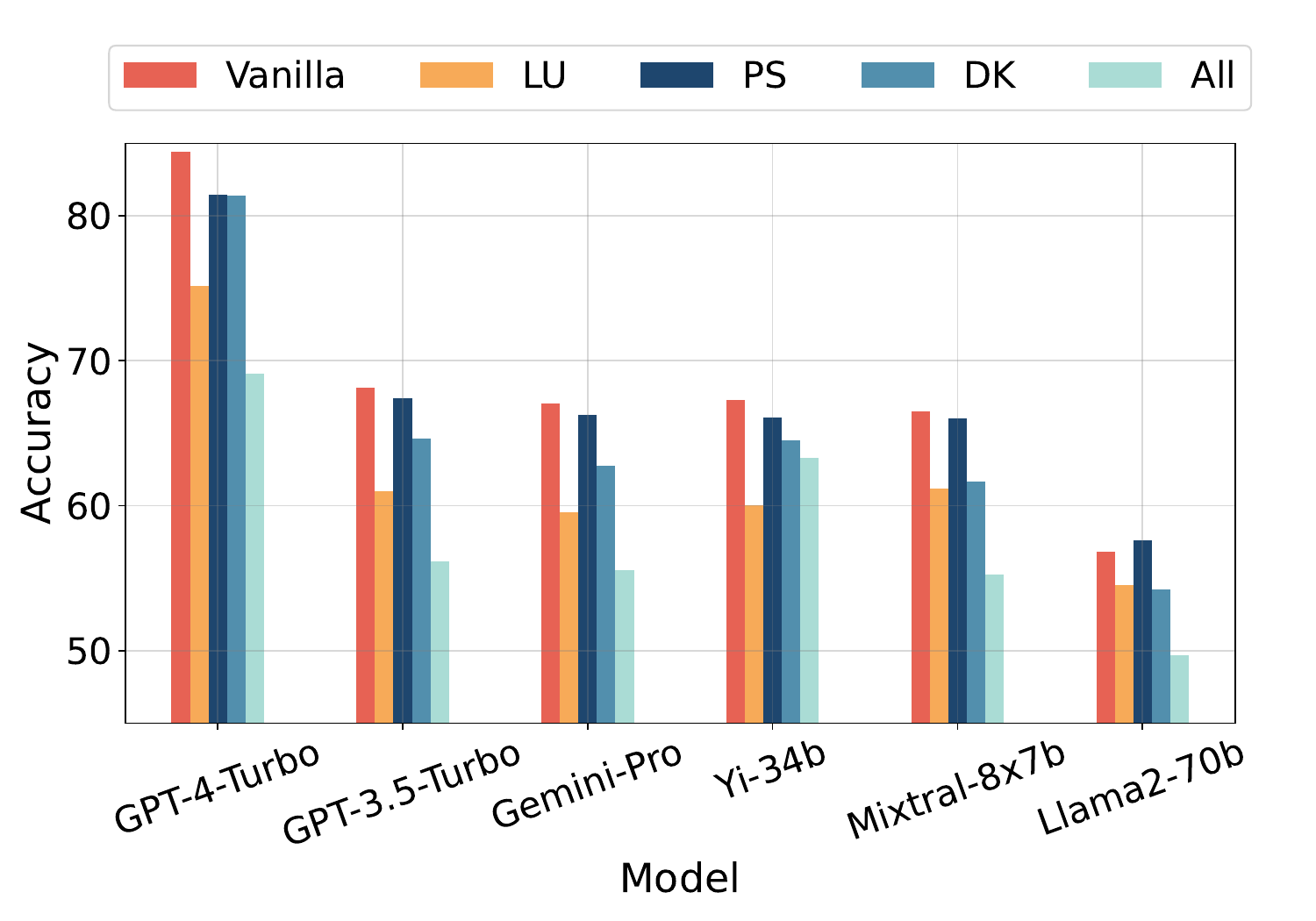}
    \vspace{-.2in}
    \caption{Performance of different \llms on vanilla \mmlu testset and our probing benchmarks based on the \mmlu. LU, PS, and DK denote the evaluation sets to evaluate language understanding, problem solving, and domain knowledge ability, respectively.}
    \vspace{-.2in}
    \label{fig-probe-judge}
\end{figure}

Recently, \citet{zhu2023dyval} proposed DyVal to dynamically generate test samples based on the graph structure to combat data contamination.
\citet{fan2023nphardeval} introduced NPHardEval, which generates new evaluation samples for NP-hard math problems and updates the evaluation set monthly.
Both are designed to evaluate reasoning tasks and cannot be easily extended to other popular natural language tasks~\citep{mmlu,clark2018think}.
For fine-grained performance analysis, \citet{burnell2023revealing} inspected the correlation between different tasks using HELM benchmark results~\citep{liang2022holistic} and concluded that the performance of \llms is not monolithic but exhibits variance between different aspects, such as reasoning and understanding.
Therefore, developing a dynamic evaluation protocol to support diverse tasks and multifaceted ability analysis remains a challenge. 

\begin{figure*}[t!]
    \centering
    \includegraphics[width=\textwidth]{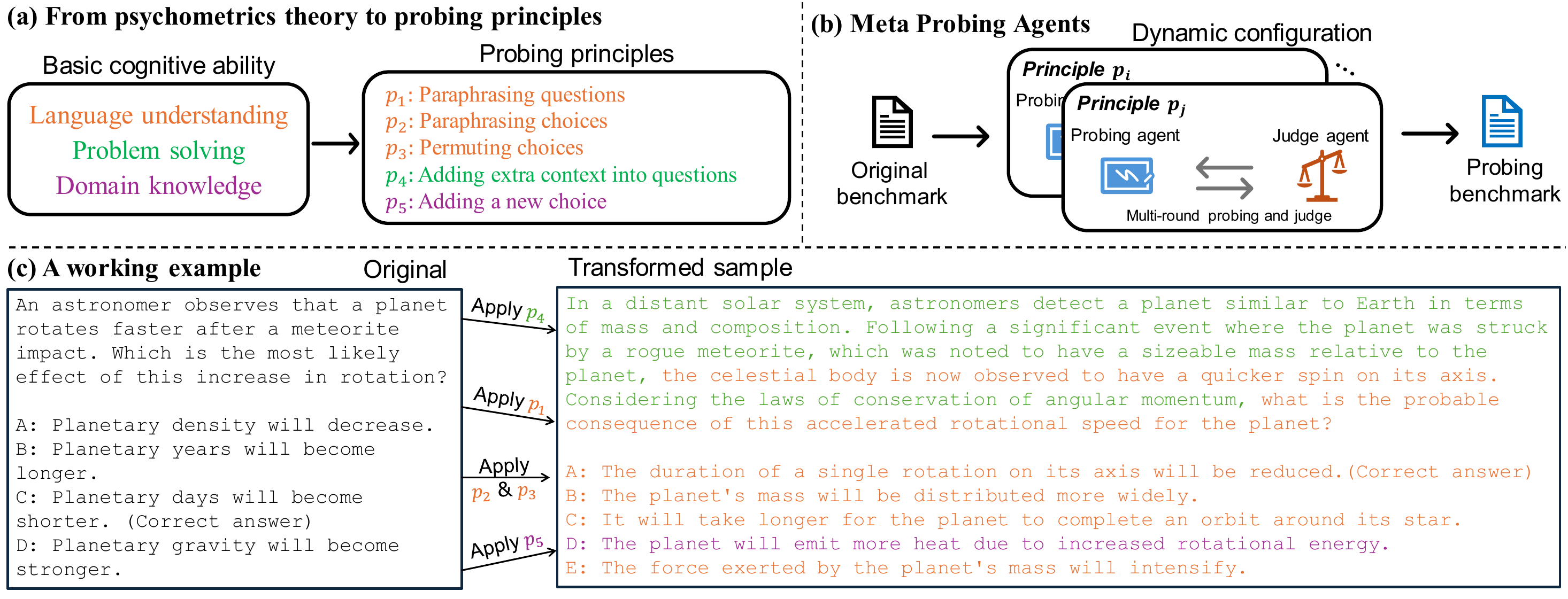}
    \vspace{-.2in}
    \caption{Inspired by psychometric theory on the three basic cognitive abilities, our Meta Probing Agent (\method) designs corresponding principles that transforms original benchmarks into a new one. These principles can be flexibly combined to create various probing benchmarks for multifaceted analysis. Subfigure (c) shows how \method generates the new sample given an existing sample from ARC-C.}
    \vspace{-.1in}
    \label{fig-pipeline}
\end{figure*}

In this paper, we propose \textbf{M}eta \textbf{P}robing \textbf{A}gents (\method), a dynamic and flexible evaluation protocol for \llms based on agents.
\method bridges the gap between psychometrics and \llms evaluation by designing  principles to dynamically generate new questions (\cref{fig-pipeline}).
The principles correspond to the three basic cognitive abilities of psychometric theory~\citep{burnell2023revealing}: language understanding, problem solving, and domain knowledge.
Therefore, \method supports both dynamic evaluation sample generation and multifaceted ability analysis.
Specifically, instead of relying on the graph structure to generate samples like DyVal, \method uses LLM-based agents to automatically transform existing problems into new ones, which are more flexible and support diverse tasks.
We define them as \emph{probing} agents, aiming to uncover the underlying knowledge in a question.
\method further utilizes a \emph{judge} agent \citep{dubois2023alpacafarm, li2023alpacaeval} to validate the generated evaluation samples. This adversarial manner ensures that the new samples maintain consistency and relevance compared to their original counterparts.
Furthermore, \method can dynamically combine various probing principles for multifaceted evaluations of the abilities.
This modular design affords researchers the flexibility to apply any combination of principles, aligning the evaluation scope with their investigative focus, and mirroring the multifaceted nature of human cognition.

We used \method to generate new evaluation sets based on popular benchmarks: \mmlu~\citep{mmlu}, \gsm~\citep{cobbe2021training}, \bbh~\citep{suzgun2022challenging}, and \arc~\citep{clark2018think}.
Then, we conducted extensive evaluations and analysis on popular \llms: \gptfour, \gptthree, \gemini~\citep{team2023gemini}, \llama~\citep{llama}, \yi~\citep{yi}, and \mixtral~\citep{mixtral}.
The takeaways of our key findings are as follows:
\begin{itemize}[leftmargin=1em]
\setlength\itemsep{0em}
    \item The performance of \llms on our dynamic benchmarks degraded significantly, implying potential data contamination on current benchmarks (\cref{fig-probe-judge} \& \S \ref{sec-exp-result}). Prompt engineering can only bring marginal improvements (\S \ref{sec-exp-prompt-engineer}).
    \item All \llms exhibited performance decreases by dynamically combining different principles (\S \ref{sec-analysis-complexity});
    \item Our multifaceted analysis demonstrated strong correlations between the three basic abilities, where language understanding and problem solving abilities have the strongest correlation (\S \ref{sec-analysis-correlation}).
    \item We observed an implicit ``Matthew effect'' between model size and correlations of the abilities: larger models tend to have stronger correlations (\S \ref{sec-analysis-modelsize}).
    \item \llms exhibited various error patterns in our fine-grained analysis pertaining to the three basic abilities (\S \ref{sec-analysis-error}).
    
    \item \method can be used as a data augmentation approach to improve the performance of \llms (\S \ref{sec-finetune}).
\end{itemize}

The contributions of this paper are summarized as follows:
\begin{itemize}[leftmargin=1em]
\setlength\itemsep{0em}
    \item \textbf{A psychometrics-inspired dynamic evaluation Protocol.} \method is general and flexible to mitigate data contamination and facilitate multifaceted analysis.
    \item \textbf{Comprehensive analysis of the basic abilities of \llms.} Our modular design allows for the dynamic combination of the three basic cognitive abilities, providing systematic evaluation and analysis for future research.
\end{itemize}

\section{Related Work}
\label{sec-related}

\textbf{\llms Evaluation \& Data Contamination.}
Various benchmarks have been introduced to assess \llms \citep{mmlu, alpaca_eval, zhong2023agieval, leaderboard,chang2023survey}, including:
(1) Human-centric evaluation, typified by AdaVision \citep{gao2022adaptive} and AdaTest \citep{ribeiro2022adaptive} that emphasize human-driven feedback.
(2) Crowd-sourcing, e.g., DynaBench \citep{kiela2021dynabench} and DynaBoard \citep{ma2021dynaboard}, which prioritizes diverse and comprehensive evaluations through crowd-sourced tests.
(3) More challenging tasks such as HELM~\citep{liang2022holistic}, DeepTest \citep{tian2018deeptest} and CheckList \citep{ribeiro2020beyond}, create custom tests, while platforms such as Big-Bench \citep{srivastava2023beyond} aim to challenge \llms with specialized tasks.

Recent research has highlighted the significant issue of data contamination in \llms \citep{lovin2023gpt4, stochasticparrot, kocon2023chatgpt, li2023open, zhou2023don}.
In particular, the reports of GPT-4~\citep{gpt4}, LLama~\citep{llama}, and Skywork LLM \citep{wei2023skywork} have acknowledged this phenomenon.
Several researchers \citep{golchin2023data, golchin2023time, oren2023proving, yang2023rethinking, oren2023proving} developed methods to detect data contamination. \citet{zhu2023dyval, lei2023s3eval, fan2023nphardeval} introduced dynamic evaluation strategies via different algorithms to reduce data contamination.
\method significantly differs from them in the sample generation mechanism and the support for multifaceted analysis.

\textbf{\llms as Autonomous Agents.}
The adoption of \llms as autonomous agents for task completion has recently gained popularity, such as AutoGPT~\citep{AutoGPT} and MetaGPT~\citep{hong2023metagpt}, which have advanced our understanding of collaborative and planning abilities in \llms. Another growing trend is the use of \llms as judges~\citep{dubois2023alpacafarm, li2023alpacaeval, fernandes2023devil, bai2023benchmarking, pandalm2024}, where LLMs assess the output of other LLMs. Furthermore, there is a burgeoning interest in leveraging LLMs to enhance training datasets~\citep{wang2022self, yu2023metamath, yuan2024self, liu2024augmenting} and aligning \llms' outputs with specific goals or values~\citep{burns2023weak}.
Our work differs from them in designing psychometrically inspired principles for sample generation and support for multifaceted analysis.

\section{Method}

\subsection{Overview}

There are two critical challenges in designing a dynamic evaluation protocol.
First, there is no general principle to guide the evaluation sample generation process for diverse tasks such as knowledge, language understanding, reasoning, and mathematics.
Existing literature like DyVal~\citep{zhu2023dyval} and NPHardEval~\citep{fan2023nphardeval} adopted manually designed principles to generate samples for specific tasks and cannot easily extend to other scenarios.
Second, the principle of generating evaluation samples should be fine-grained yet atomic enough to analyze the multifaceted capabilities of \llms.
Our hope is that the evaluation should reflect the primitive abilities, and more fine-grained analysis of their correlations can be conducted.

In this work, we take inspiration from psychometrics~\citep{raykov2011introduction,burnell2023revealing} to generate evaluation samples using \llms as agents.
Specifically, instead of relying on specific rules like DyVal~\citep{zhu2023dyval}, we employ \llms as agents to automatically generate new problems based on the given evaluation samples.
This agent-based evaluation design can potentially fit most tasks.
More importantly, psychometric theory categorizes cognitive abilities into three basic ones: \emph{language understanding} to comprehend and generate texts, \emph{problem-solving} to deal with complex problems, and \emph{domain knowledge}.
Therefore, as shown in \cref{fig-pipeline}(a), we follow these criterion to design principles that not only change the problems but also assist in multifaceted analysis.

\begin{table*}[t!]
\caption{The prompts for different principles based on the three cognitive abilities.}
\label{table: principle prompt}
\centering  
\resizebox{\textwidth}{!}{
\begin{tabular}{cccc}
\toprule

\textbf{Principle} & \textbf{Ability} &\textbf{Agent Type} & \textbf{Prompt} \\

\midrule

\multirow{4}{*}{\makecell{Paraphrasing\\Question}} & \multirow{4}{*}{\makecell{Language\\Understanding}} & \multirow{2}{*}{Probing} & \prompt{\makecell[tl]{I plan to paraphrase the question to present the same concept in a different way.\\Please assist me in paraphrasing the question.}} \\

\cmidrule(lr){3-4}

& & \multirow{2}{*}{Judge} & \prompt{\makecell[tl]{Your task is to analyze both the original question and the revised question and\\determine if they are effectively assessing the same concept or knowledge area.}} \\

\midrule
\multirow{6}{*}{\makecell{Paraphrasing\\Choices}} & \multirow{6}{*}{\makecell{Language\\Understanding}} & \multirow{3}{*}{Probing} & \prompt{\makecell[tl]{I plan to paraphrase the choices: each choice should be paraphrased to reflect the\\same concept or idea as the original. The essence and meaning must be preserved. If\\a choice cannot be paraphrased without changing its meaning, it should be kept unchanged.}} \\

\cmidrule(lr){3-4}

& & \multirow{3}{*}{Judge} & \prompt{\makecell[tl]{Your task is to analyze the paraphrased choices in the context of the question and\\determine if the paraphrased choices (A, B, C, D) still reflect the original meaning of\\their respective original choices.
}} \\

\midrule

\multirow{1}{*}{Permuting Choices} & \makecell{Language\\Understanding} & - & (This principle does not require agents and can be implemented directly in the code) \\

\midrule
\multirow{6}{*}{\makecell{Adding Extra Context\\into Question}} & \multirow{6}{*}{\makecell{Problem\\Solving}} & \multirow{3}{*}{Probing} & \prompt{\makecell[tl]{I plan to add non-essential context to the question: introducing context or details \\ to the question that are relevant but not directly helpful in answering it.\\ The context can be put at the beginning, middle, or end of the question.}} \\

\cmidrule(lr){3-4}

& & \multirow{2}{*}{Judge} & \prompt{\makecell[tl]{Your task is to analyze both the original question and the revised question and\\determine if they are effectively assessing the same concept or knowledge area.}} \\

\midrule
\multirow{6}{*}{\makecell{Adding\\A New Choice}} &  \multirow{6}{*}{\makecell{Domain\\Knowledge}} & \multirow{3}{*}{Probing} & \prompt{\makecell[tl]{I plan to keep the choices A,B,C,D unchanged, and introduce an additional relevant\\choice E that is related to the topic but doesn't provide another correct answer. \\This choice should be plausible but clearly not correct in the context of the question.}}  \\

\cmidrule(lr){3-4}

& & \multirow{3}{*}{Judge} & \prompt{\makecell[tl]{Your task is to analyze the paraphrased choices in the context of the question and\\determine whether the new choice (E) is relevant to the question but does not\\provide an alternative correct answer.
}} \\
\bottomrule
\end{tabular}
}
\end{table*}

Our approach, termed as \emph{meta probing agents} (\method), is illustrated in \cref{fig-pipeline}(b).
Given an original evaluation sample from existing benchmarks, \method aims to evaluate the ability of \llms by creating new problems following psychometric theory.
We create a set of principles that involve two agents: the \emph{probing} agent and the \emph{judge} agent.
The probing agent aims to investigate the knowledge of a given question $Q$ and return a new one according to a principle $p_i$, and the judge agent is responsible for the validity and consistency check based on the original question.
Their interaction features a feedback mechanism: if the judge agent determines that the new question lacks consistency (`No'), the probing agent is prompted again to generate another version of the question. Conversely, if the answer is `Yes', the question is deemed to have passed the consistency check.

The dynamic nature of \method lies in two aspects: the dynamic generation of evaluation samples and the dynamic combination of principles.
Such a dynamic combination allows for multifaceted analysis of \llms' abilities, thus providing more insight for future research.
It can also be seen as granting problems with flexible complexities~\citep{zhu2023dyval} for a comprehensive evaluation.
Now, we present the details of the agents and the psychometric principles.


\subsection{Probing Agent}
The probing agent aims to transform a given question into a new one to assess \llms' ability required by a question $Q$. 
The probing is guided by principles inspired by psychometrics, which are encapsulated within a carefully crafted prompt.
Unlike generating training samples, one of the most important criteria in probing agents is to ensure that the generated questions maintain the core essence of the original while presenting a different perspective, thus maintaining its correctness.
An example is shown in \cref{fig-pipeline}(c), where a sample of \arc is transformed into a new one by applying different principles.
The prompts in our experiments incorporate directives that guide the agent towards creating semantically similar but structurally different questions.


\subsection{Judge Agent}
Although we have explicitly restricted the probing agent to maintain the integrity of the original question, there are cases where probing agents unintentionally change the meaning. 
Thus, the judge agent is designed to provide a clear and unambiguous assessment of whether the generated question maintains the integrity of the original intent and informational content.
Its prompt is designed to direct \llms to compare the original with the rephrased questions, ensuring the preservation of the essence and factual accuracy.

Specifically, unlike traditional evaluation methods that may use a variety of metrics, the judge agent operates in an adversarial manner via a binary response system through prompts. It simply returns a `Yes' or `No' verdict, indicating whether the new question maintains consistency with the original.
In this prompt, the judge agent is required to analyze the essence of both the original and rephrased questions. Its goal is not merely to identify superficial similarities or differences in wording, but to delve deeper into whether both versions of the question are aligned in terms of the concept or knowledge area they are assessing.

\paragraph{Human Verification} For each ability (language understanding, problem solving, and domain knowledge), we randomly selected $500$ samples from the MMLU dataset and $100$ samples from the ARC-c dataset, totaling $1,800$ questions. $30$ human experts (with bachelor or higher degree) are divided into 3 groups, each with 10 person. They were asked to judge the following two questions: (1) whether the original and paraphrased questions were equivalent; (2) if the answers to the probing questions remained correct.
The evaluation required a simple `Yes' or `No' answer.

The positive results shown in Appendix~\cref{tb: appendix-human} from our human evaluation with an overall accuracy rate of $94$\% and $97$\% for each question, underscoring the effectiveness of our methodology. The following table provides a detailed breakdown of the evaluation outcomes, showcasing the high level of confidence in the equivalence and correctness of our probing questions across the three abilities.

\subsection{Psychometric principles}
\label{sec-method-psychometric}

Psychometric principles guide how we probe the understanding of a question. 
Inspired by psychometric theory~\citep{raykov2011introduction}, we aim to evaluate three basic abilities of \llms: language understanding, problem solving, and domain knowledge.
We have identified five key principles that correspond to these categories, as shown in \cref{table: principle prompt}.
These principles incorporate both a probing agent and a judge agent, as previously discussed, with their functions differing according to the specific principle applied.

\subsubsection{Language Understanding}
Language understanding assesses the ability to process, interpret, and generate texts.
To evaluate this ability, we focus on how well \llms can grasp the underlying meaning of various linguistic expressions and maintain its integrity when presented in different forms.
We design three principles to evaluate this ability:
\begin{itemize}[leftmargin=1em]
\setlength\itemsep{0em}
    \item \textbf{Principle 1: Paraphrasing Questions.}
    It focuses on altering the phrasing of a question while retaining its core concept. This is achieved via a prompt that guides the probing agent to restructure the question without changing its underlying meaning. The new questions challenge \llms in understanding to ensure that they grasp the essence of the question beyond surface-level recognition.
    
    \item \textbf{Principle 2: Paraphrasing Choices.}
    It is similar to the first one, but applies to choices in a multiple-choice format. It involves rephrasing the options provided in a way that maintains their original intent and meaning.\footnote{Most benchmarks adopt the QA style. For non-QA benchmarks such as \gsm, the choice-related principles do not apply.}
    \item \textbf{Principle 3: Permuting Choices} \citep{zong2023fool}. This principle simply involves rearranging the order of the choices in a multiple-choice question. It determines if the model's understanding is influenced by the position of the correct answer. As it can be achieved through coding, specific prompts are not required for this principle.
\end{itemize}

\subsubsection{Problem Solving}
Problem solving refers to the ability to analyze, deduce, and derive answers.
It involves critical thinking, distinguishing relevant from irrelevant data, and applying knowledge to new situations. Principles under this category test the model's ability in navigating complex, often nuanced scenarios, and its proficiency in delivering solutions.
Note that we cannot create completely new problems by the agent, since their correctness cannot be guaranteed.
Therefore, we design one general principle:
\begin{itemize}[leftmargin=1em]
\setlength\itemsep{0em}
    \item \textbf{Principle 4: Adding Extra Context into Questions.} It aims to introduce additional, non-essential context to the question, which is relevant to the topic, but does not directly aid in answering the question. The prompt guides the probing agent to seamlessly integrate extra context into the original question. The new questions assess whether \llms can filter out extraneous information and focus on the key elements to solve the problem.
\end{itemize}

\subsubsection{Domain Knowledge}
Domain knowledge refers to the depth and accuracy of the model's knowledge in specific areas. It is crucial not only to have a broad understanding of general concepts, but also to possess detailed, nuanced knowledge.
It tests the model's expertise in various domains, its ability to differentiate between closely related concepts, and to apply this knowledge appropriately in context-specific scenarios.
\begin{itemize}[leftmargin=1em]
\setlength\itemsep{0em}
    \item \textbf{Principle 5: Adding A New Choice.} It focuses on supplementing existing choices with an additional one. The new choice is relevant to this question, but is not a correct answer, which relies on domain knowledge to exclude.
\end{itemize}

\emph{Remark:}
\method is not limited to these five principles and more can be added easily through our framework.


\section{Experiments}
\subsection{Experimental Setup}

\textbf{Tasks and Datasets.}
We selected four popular datasets for evaluation: \mmlu~\citep{mmlu}, ARC-Challenge (\arc)~\citep{clark2018think}, \gsm~\citep{cobbe2021training}, and BigBench-Hard (\bbh)~\citep{suzgun2022challenging, srivastava2022beyond}, encompassing a broad spectrum of computational challenges ranging from knowledge-intensive understanding to complex mathematical and logical reasoning tasks. We adopted only three hard tasks from BBH: Formal Fallacies, Object Counting, and Temporal Sequences. We used their test sets to generate new evaluation samples. The detailed introduction is given in \cref{append-dataset}.

\textbf{Evaluated \llms.}
We evaluated three proprietary \llms: \gptfour~\citep{gpt4}, \gptthree~\citep{chatgpt}, and \gemini~\citep{team2023gemini}, and three strong open-sourced models: \llama~\citep{llama}, \yi~\citep{yi}, and \mixtral~\citep{mixtral}. To ensure a standardized comparison, we set the generation temperature to $0$ for all models, with the generation length as $1000$ tokens.\footnote{For \gemini on \mmlu dataset, we set the temperature to $0.7$ and omitted some evaluation samples (around 20 samples) to avoid response failures due to Google's safety constraints.}

\textbf{Agent \llms in \method.}
We utilized \gptfour as probing and judging agents, with temperatures of $0.7$ and $0$, respectively. The maximum token generation for each agent is set as $1000$. While \gptfour serves as the main agents, we also explored the potential to integrate other \llms such as \gptthree and \gemini as agents in later experiments in \cref{weak-llm}. 
The evaluation prompts are detailed in the Appendix~\ref{append-eval-prompts}. Our primary evaluation metric is accuracy.

A bitter reality is that currently, only GPT-4 are capable to generate questions and judge the quality of generated questions. We believe as the field progresses, more cheaper models (such as Claude 3 and Gemini) will become capable of fulfilling both probing and judging roles, thereby reducing dependency on any single model's API and enhancing the scalability. 

\subsection{Main Results}
\label{sec-exp-result}

\begin{table*}[t!]
\caption{The performance of different \llms on vanilla benchmarks and our probing benchmarks.}
\label{table: main results}
\centering
\resizebox{1\textwidth}{!}{
\begin{tabular}{c|ccc|ccc|ccc|ccc|ccc|ccc}
\toprule
Model & \multicolumn{3}{c|}{GPT-4-Turbo} & \multicolumn{3}{c|}{GPT-3.5-Turbo} & \multicolumn{3}{c|}{Gemini Pro} & \multicolumn{3}{c|}{Yi-34b} & \multicolumn{3}{c|}{Mixtral-8x7b} & \multicolumn{3}{c}{Llama2-70b-chat} \\
Dataset & \multicolumn{1}{c}{Vanilla} & \multicolumn{1}{c}{Ours} & $\Delta$ & \multicolumn{1}{c}{Vanilla} & \multicolumn{1}{c}{Ours} & $\Delta$ & \multicolumn{1}{c}{Vanilla} & \multicolumn{1}{c}{Ours} & $\Delta$ & \multicolumn{1}{c}{Vanilla} & \multicolumn{1}{c}{Ours} & $\Delta$ & \multicolumn{1}{c}{Vanilla} & \multicolumn{1}{c}{Ours} & $\Delta$ & \multicolumn{1}{c}{Vanilla} & \multicolumn{1}{c}{Ours} & $\Delta$ \\ \midrule
MMLU & \multicolumn{1}{c}{84.40} & \multicolumn{1}{c}{68.86} & -15.54 & \multicolumn{1}{c}{68.12} & \multicolumn{1}{c}{56.15} & -11.97 & \multicolumn{1}{c}{67.04} & \multicolumn{1}{c}{55.55} & -11.49 & \multicolumn{1}{c}{67.31} & \multicolumn{1}{c}{63.30} & -4.01 & \multicolumn{1}{c}{66.49} & \multicolumn{1}{c}{55.24} & -11.25 & \multicolumn{1}{c}{56.85} & \multicolumn{1}{c}{49.70} & -7.15 \\ 
\gsm & \multicolumn{1}{c}{95.22} & \multicolumn{1}{c}{88.50} & -6.72 & \multicolumn{1}{c}{77.71} & \multicolumn{1}{c}{71.54} & -6.17 & \multicolumn{1}{c}{22.97} & \multicolumn{1}{c}{20.39} & -2.58 & \multicolumn{1}{c}{73.54} & \multicolumn{1}{c}{68.54} & -5.00 & \multicolumn{1}{c}{61.56} & \multicolumn{1}{c}{47.18} & -14.38 & \multicolumn{1}{c}{52.92} & \multicolumn{1}{c}{51.50} & -1.42 \\ 
ARC-C & \multicolumn{1}{c}{96.16} & \multicolumn{1}{c}{84.67} & -11.49 & \multicolumn{1}{c}{85.41} & \multicolumn{1}{c}{74.60} & -10.81 & \multicolumn{1}{c}{86.18} & \multicolumn{1}{c}{75.91} & -10.27 & \multicolumn{1}{c}{86.78} & \multicolumn{1}{c}{74.03} & -12.75 & \multicolumn{1}{c}{84.47} & \multicolumn{1}{c}{70.36} & -14.11 & \multicolumn{1}{c}{73.55} & \multicolumn{1}{c}{64.19} & -9.36 \\ 
BBH (partial) & \multicolumn{1}{c}{88.53} & \multicolumn{1}{c}{87.78} & -0.75 & \multicolumn{1}{c}{54.67} & \multicolumn{1}{c}{49.73} & -4.94 & \multicolumn{1}{c}{65.47} & \multicolumn{1}{c}{60.00} & -5.47 & \multicolumn{1}{c}{55.47} & \multicolumn{1}{c}{52.49} & -2.98 & \multicolumn{1}{c}{53.47} & \multicolumn{1}{c}{40.53} & -12.94 & \multicolumn{1}{c}{38.53} & \multicolumn{1}{c}{38.22} & -0.31 \\ \bottomrule
\end{tabular}
}
\vspace{-.1in}
\end{table*}

In this part, we applied all five principles to generate new evaluation samples for \mmlu and \arc. For \gsm and \bbh that do not have multiple choices, we restricted our probing to Principle $1$ and $4$.
\cref{sec-append-examples} shows some examples generated by \method.
We repeated the evaluation three times to reduce randomness.
\cref{table: main results} presented the test accuracy of different \llms on the original and our \method benchmarks.
Standard deviation are mostly around $1$ (\cref{table: std}), indicating the robustness of the benchmark.
As can be seen, all \llms exhibited performance degradation on our probing benchmarks. Although \gptfour demonstrated the strongest data contamination problem, it remains the strongest model.
For \mmlu, \gptfour performed $15.7$\% worse than the original benchmark. Furthermore, a notable performance decline is evident in the case of \mmlu and \arc, which is significantly more pronounced than that observed in \gsm and \bbh. This suggests that \llms may encounter the memorization of knowledge-based benchmarks, resulting in substantial performance degradation for evaluation on our benchmarks.

\begin{figure}[t!]
    \centering
    \includegraphics[width=0.45\textwidth]{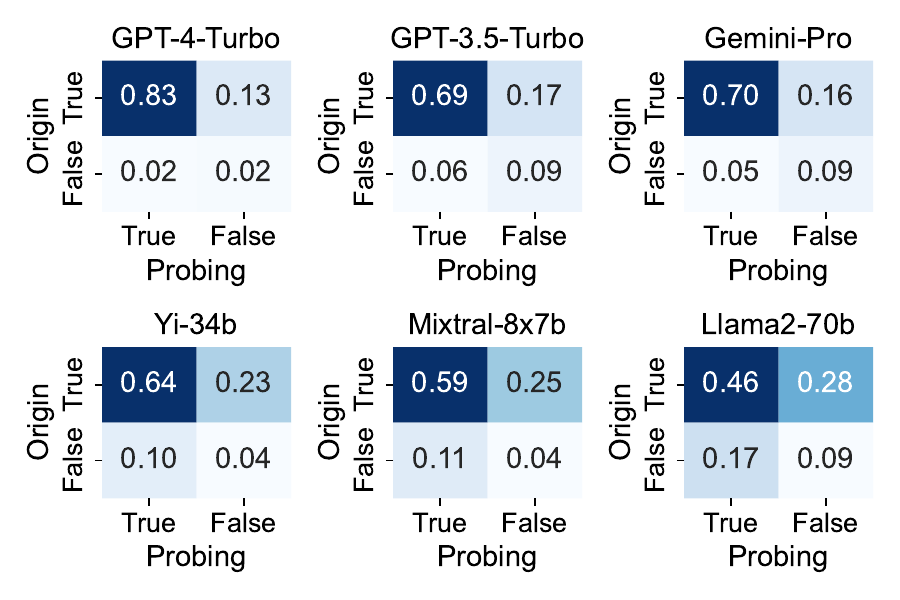}
    \vspace{-.15in}
    \caption{The confusion matrix of original benchmarks and probing benchmarks on \arc dataset.}
    \label{fig-conf-mat}
    \vspace{-.15in}
\end{figure}

We also presented a confusion matrix for analysis as shown in \cref{fig-conf-mat}. 
The matrix categorizes the responses into four distinct segments, with four categories to evaluate the model responses: OT (Original True), PT (Probing True), OF (Original False), and PF (Probing False). 
A notable observation is the high frequency of `OT/PF' instances, which suggests a potential data contamination to specific dataset characteristics.  
Furthermore, the frequency of open-source models is markedly higher than that of proprietary models like GPT-4. This discrepancy indicates that open-source models might be more susceptible to data contamination.

\subsection{Effect of Different Probing Principles}
\label{sec-exp-diff-principle}

We further studied the effects of each modular probing principle. To this end, we established a baseline where the combined effect of all principles leads to a decrease in performance, normalized to a value of ``1''. This approach enables us to compare the relative effectiveness of each principle in isolation. The performance decrement for each principle, when applied independently, was evaluated and compared with this baseline. The Relative effectiveness (RE) is computed as:
$    \text{RE} = \frac{\text{Acc}_{p_i} - \text{Acc}}{\text{Acc}_{p_{all}} - \text{Acc}},
$
where $\text{Acc}_{p_i}$ is the accuracy when only apply principle $p_i$ to \method, $\text{Acc}_{p_{\text{all}}}$ denotes the accuracy of \method when all principles are applied. $\text{Acc}$ is the accuracy on the original benchmark.

\begin{figure}[t!]
    \centering
    \includegraphics[width=0.48\textwidth]{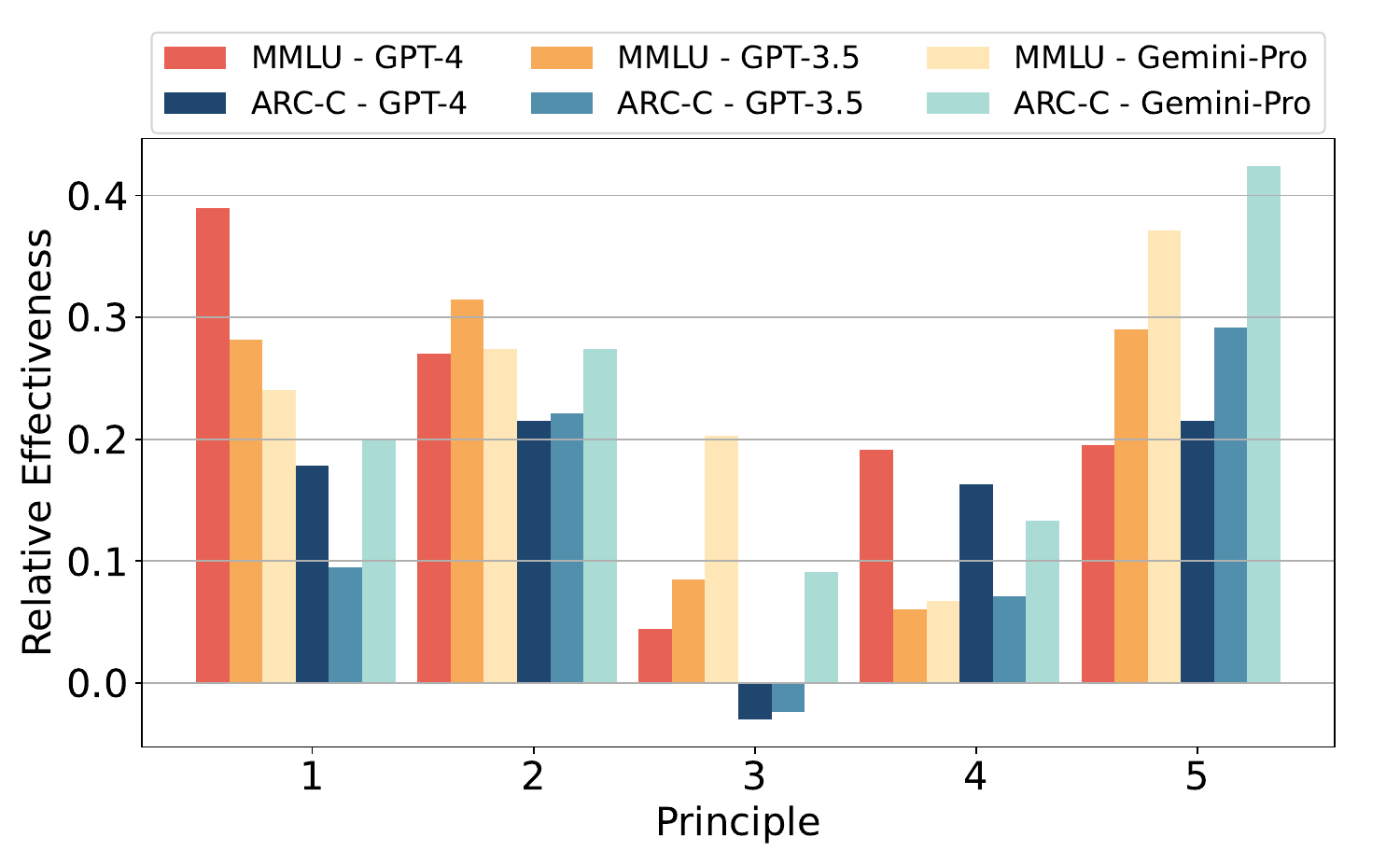}
    \vspace{-.2in}
    \caption{The relative effectiveness of different principles on \mmlu and \arc dataset.}
    \label{fig-ablation-different-principle}
    \vspace{-.2in}
\end{figure}

The results on \mmlu and \arc datasets are detailed in \cref{fig-ablation-different-principle} and \cref{sec-apped-ablation-principles}. It can be observed that, principle $1,2$ and $5$ are the most effective principles. While principle $3$, which randomly permute choices, are less effective. For \gptfour and \gptthree on \arc dataset, it can intriguingly increase the performance.

\subsection{Ablation Study on Prompt Engineering}
\label{sec-exp-prompt-engineer}

We explored the efficacy of prompt engineering techniques, specifically Chain-of-Thought (CoT)~\citep{wei2022chain} and In-Context Learning (ICL)~\citep{brown2020language}.
For ICL, we selected five examples from the corresponding training set as few-shot examples. We also applied \method to these five examples, creating a transformed version of few-shot examples. We refer to the use of ICL with the original training examples as ICL$_o$ and the use of ICL with the transformed examples as ICL$_t$.

As can be observed in \cref{table: pe results}, neither CoT nor ICL can effectively boost the performance of \llms on our probing benchmarks. 
Performance enhancements are varied across models and datasets when CoT and ICL are applied. For instance, \gptfour showed a modest increase in accuracy on the \gsm dataset with ICL, improving from $88.50$ to $89.39$. 
In contrast, \gptthree's performance slightly decreased under the same conditions. 
Note that the significant performance gain observed for \gemini on the \gsm dataset when using ICL is attributed to its initial lower effectiveness in a zero-shot context.

\begin{table}[t!]
\centering
\caption{Results of different prompt engineering techniques for \gsm and \arc dataset, with the highest and second-highest accuracies highlighted in bold and underlined, respectively.}
\label{table: pe results}
\resizebox{0.45\textwidth}{!}{
\begin{tabular}{cccccc}
\toprule
Dataset    & Model     & Original       & CoT            & ICL$_o$            & ICL$_t$           \\
\midrule
\multirow{3}{*}{GSM8K}         & \gptfour  & 88.50          & {\ul 89.31}    & \textbf{89.39} & 88.78          \\
                               & \gptthree & \textbf{71.54} & {\ul 70.58}    & 65.73          & 64.90          \\
                               & \gemini   & 20.39          & 24.49          & \textbf{75.28} & {\ul 73.01}    \\
\midrule
\multirow{3}{*}{ARC-C} & \gptfour  & 84.67          & 82.85          & {\ul 85.32}    & \textbf{85.67} \\
                               & \gptthree & {\ul 74.60}    & \textbf{75.94} & 74.32          & 74.49          \\
                               & \gemini   & 75.91          & 76.02 & {\ul 76.71} & \textbf{79.10} \\
\bottomrule
\end{tabular}
}
\vspace{-.1in}
\end{table}

\subsection{Albation Study on Data Contamination}

We collected $30$ novel reasoning questions from the experts (the same experts in the human evaluation in the general response), and examined the performance of the original questions and the probing questions. The GPT-4's accuracies of the original questions and the probing questions are both $60$\%. The OT/PF ratio is $3$\%, which is much lower than those in common public benchmarks, indicating that the newly generated benchmarks are not likely to be memorized compared to the public benchmarks.

\subsection{Weak \llms as Probing and Judge Agents}
\label{weak-llm}

In this section, we assess the feasibility of using less advanced \llms to reduce costs of \method evaluation.
We initially configured \gptthree and \gemini as judging agents, while maintaining \gptfour as the probing agent for \gsm dataset. Through meticulous manual examination of the questions transformed by this setup, we observed a significant shortfall in the judging agents' ability to discern and exclude transformed examples whose meanings deviated from the original questions. highlighting the need for robust and capable \llms to effectively sieve out appropriately probing questions.
Subsequently, we extended our inquiry to assess the potential of employing less capable LLMs like \gptthree and \gemini as probing agents.
Using \gptfour as the judging agent, our experiments revealed that when weaker \llms served as probing agents, they often altered the essence or subtleties of the original questions, leading to misrepresentations or the introduction of unintended elements. And \gptfour struggled to consistently detect these nuanced alterations. This suggests that the sophistication of the probing agent is crucial, as even advanced judging agents like \gptfour cannot always offset the limitations of the probing agents.

\section{Multifaceted Analysis of the Basic Abilities}
\label{sec-analysis}

One advantage of \method is the support for multifaceted analysis of abilities, which is discussed in this section.

\subsection{Analysis on Benchmark Complexity}
\label{sec-analysis-complexity}

\begin{figure}[t!]
    \centering
    \subfigure[ARC-C]{
        \includegraphics[width=0.22\textwidth]{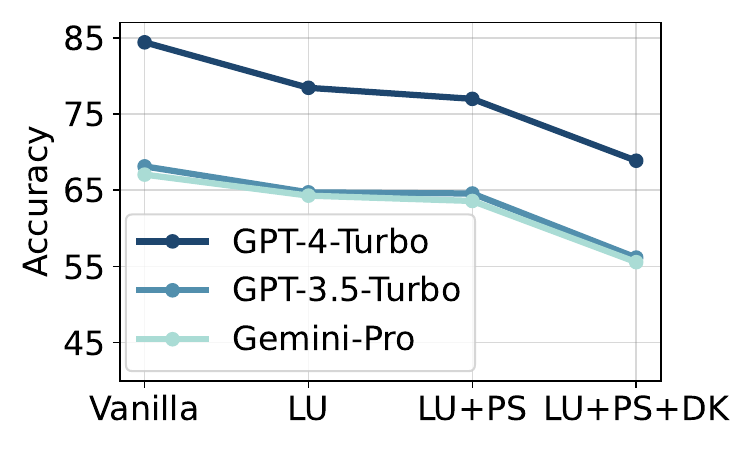}
        \label{fig: performance mmlu}
    }
    \subfigure[MMLU]{
        \includegraphics[width=0.22\textwidth]{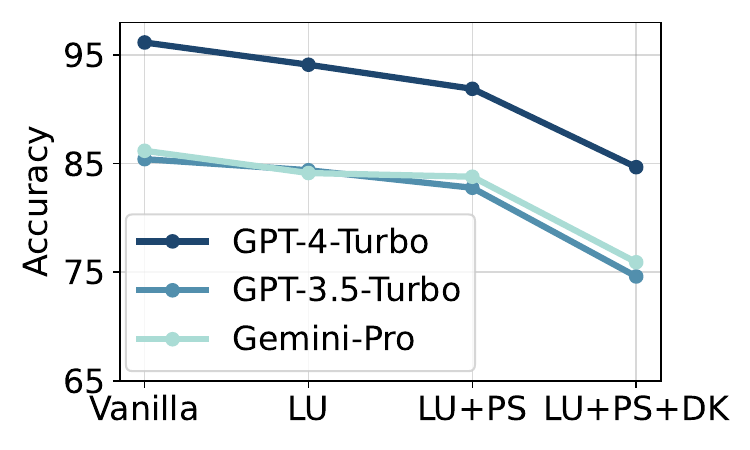}
        \label{fig: performance arc}
    }
    \vspace{-.1in}
    \caption{The accuracy of different \llms on \arc and \mmlu on different levels of probing benchmarks. LU, LU+PS, and LU+PS+DK represent probing benchmarks that applied language understanding principles, both language understanding principles and problem solving principles, and all principles, respectively.}
    \label{fig-benchmark-complexity}
    \vspace{-.1in}
\end{figure}

We first explore the impact of benchmark complexity on the \mmlu and \arc datasets, as illustrated in \cref{fig-benchmark-complexity}. We construct probing benchmarks with different levels of complexity using (1) language understanding principle $1$, (2) language understanding principle $1$ and problem-solving principle $4$, and (3) language understanding principle $1$, problem-solving principle $4$, and domain knowledge principle $5$.
In particular, as the complexity of the benchmarks increases, the performance of all \llms decreases and \gptfour consistently reaches the highest precision.
This result shows that there is still much room to improve the abilities of \llms in complex benchmarks.

\subsection{Relationship of the Basic Abilities}
\label{sec-analysis-correlation}

\begin{table}[t!]
\centering
\caption{The correlation efficient of the three basic abilities.}
\label{table: correlation}
\resizebox{0.32\textwidth}{!}{
\begin{tabular}{cccc}
\toprule
         & Pearson & Spearman & Kendall \\
\midrule
LU \& PS & 0.994   & 0.986    & 0.939   \\
LU \& DK & 0.986   & 0.972    & 0.909   \\
PS \& DK & 0.986   & 0.979    & 0.909   \\
\bottomrule
\end{tabular}
}
\vspace{-.1in}
\end{table}

To gain a deep understanding of the relationship between the abilities of language understanding (LU), problem solving (PS), and domain knowledge (DK), we constructed three probing benchmarks using the principles that belong to a certain ability and then evaluated \llms on these benchmarks.
The results are presented in \cref{tb: complexity}.
After obtaining the performance of different \llms on each probing benchmark, we then calculated the Pearson correlation coefficients~\citep{PearsonReference}, the Spearman correlation coefficients~\citep{SpearmanReference}, and the Kendall correlation coefficients~\citep{KendallReference}, the results are presented in \cref{table: correlation}.
It is evident that all abilities are highly correlated, which aligns with the findings of \citet{burnell2023revealing}.
Furthermore, it is observed that language understanding and problem solving are more relevant compared to other pairs of abilities.
This suggests that the two abilities can predict each other, which has great potential to train and improve \llms in the future.
In the future, further detailed analysis can be performed to gain a deeper insight into the basic abilities of \llms by constructing \method evaluation samples based on other existing benchmarks such as HELM~\citep{liang2022holistic}.

\subsection{Analysis on Model Size}
\label{sec-analysis-modelsize}

\begin{figure}[t!]
    \centering
    \subfigure[Model Size v.s. Accuracy]{
        \includegraphics[width=0.22\textwidth]{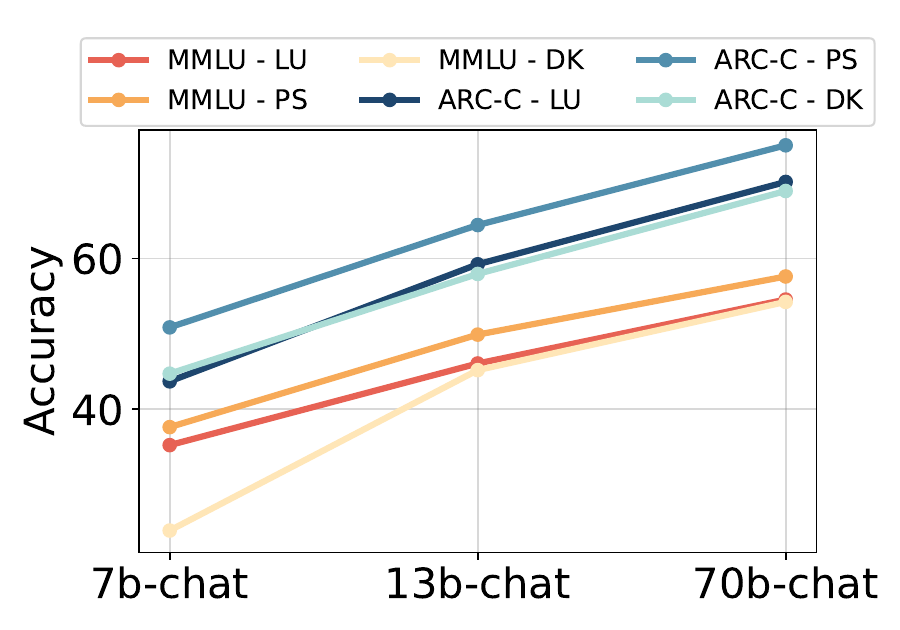}
        \label{fig: model size acc}
    }
    \subfigure[Model Size v.s. Correlation]{
        \includegraphics[width=0.22\textwidth]{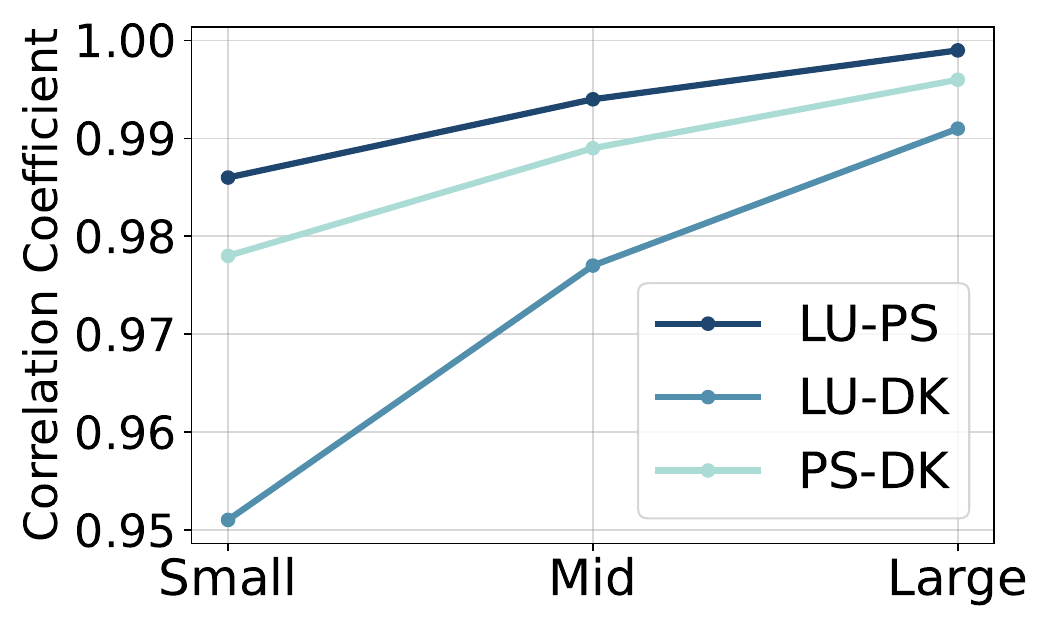}
        \label{fig: model size correlation}
    }
    \vspace{-.1in}
    \caption{(a) The correlation between the performance and model size. (b) Cross-ability correlation with model size.}
    \label{fig-model-size}
    \vspace{-.1in}
\end{figure}

We studied the influence of model size on basic abilities.
First, \cref{fig: model size acc} shows the correlation with variants of Llama2: 7b, 13b, and 70b.
It can be observed that each ability positively correlates with the model size with nearly the same slope, indicating that when the size of the model increases, all abilities are equally enhanced to improve overall performance.
Second, we explored the relationship between model size and correlations between different abilities. We roughly divided the models into three sizes: (1) small: Llama2-7b-chat, Llama2-13b-chat; (2) mid: \yi, \mixtral, \llama; and (3) \gemini, \gptthree, \gptfour. The results in \cref{fig: model size correlation} implies an implicit ``Matthew Effect''~\citep{merton1968matthew}: larger (often stronger) models tend to have stronger correlations between basic abilities.
This aligns well with existing psychological theory about the $g$ factor of general intelligence~\citep{spearman1961general}.
This finding can potentially help explain the emergent abilities of \llms~\citep{biderman2023emergent, schaeffer2023emergent} and provide insight into the evolution of \llms.

\subsection{Error Analysis}
\label{sec-analysis-error}
We conducted an in-depth analysis of \llms' failure modes in the three basic abilities. We meticulously selected $50$ instances where \gptfour correctly answered the original questions but failed in the transformed questions in the \gsm dataset. These error modes are shown below.
\begin{itemize}[leftmargin=1em]
\setlength\itemsep{0em}
    \item \textit{Language understanding.} 
    (1) Question Understanding Error: \gptfour calculates the correct answer but misinterprets the intent of the question, leading to an incorrect response. This error indicates a gap in comprehension of the question. 
    (2) Instruction Following Error: In this mode, \gptfour arrives at the correct answer but fails to present it in the required format specified in the prompt, indicating a lack of follow-up instructions.
    \item \textit{Problem solving.} Here, \gptfour understands the question correctly but errs during the calculation process, resulting in a wrong final answer.
    \item \textit{Domain knowledge.} We investigated the distribution of topic error among $57$ tasks of \mmlu in \cref{fig-error-top-20}. Notably, the \prompt{professional law} domain has the highest error rate, followed by \prompt{moral scenarios} and \prompt{professional psychology}, suggesting challenges predominantly in professional and ethical tasks.
\end{itemize}

Furthermore, we observed two possible reasons for performance degradation: ambiguity in the original questions and inconsistency between the probing and original questions. Some questions in the original dataset were found to be ambiguous (see \cref{sec-append-ambiguous-examples}). Despite this, \gptfour often provided plausible answers, suggesting potential issues with data memorization.
However, certain transformed questions deviated in meaning from their original versions, leading to inconsistencies in responses.



\begin{figure}[htbp]
    \centering
    \includegraphics[width=0.47\textwidth]{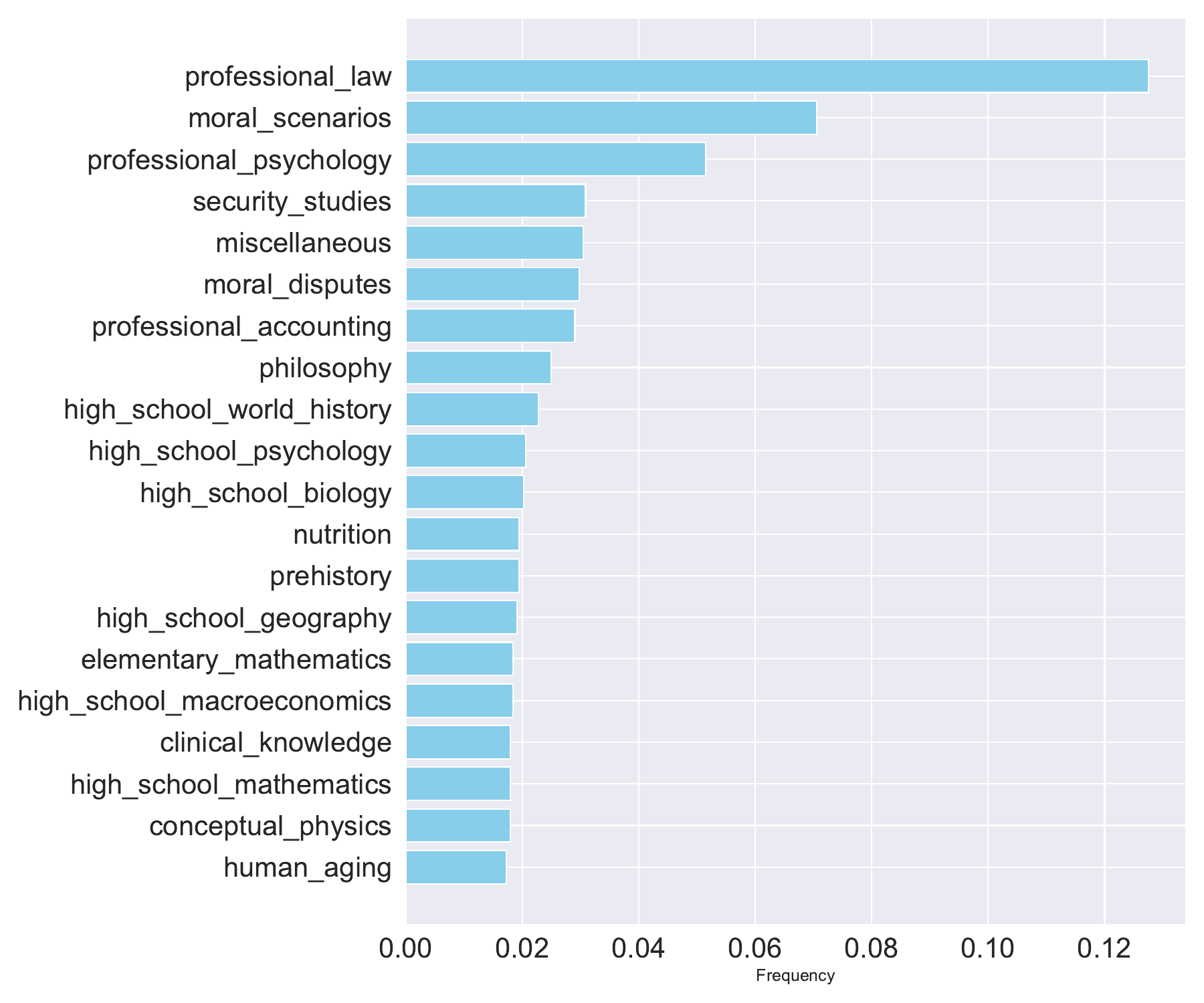}
    \caption{The bar chart of top 20 error topics and their corresponding frequencies of \gptfour on \mmlu dataset.}
    \label{fig-error-top-20}
\end{figure}


\section{\method as Data Augmentation for \llms}
\label{sec-finetune}

\begin{figure}[t!]
    \centering
    \includegraphics[width=0.4\textwidth]{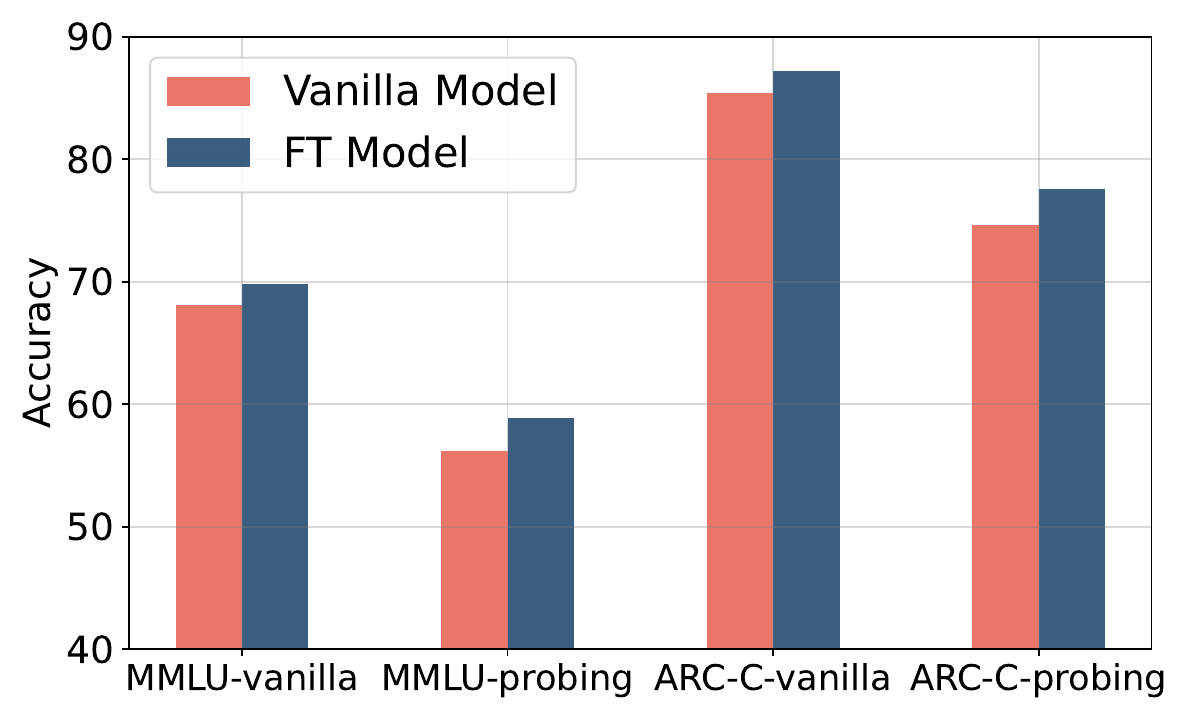}
    \vspace{-.15in}
    \caption{The performance of \gptthree and \gptthree-FT on original benchmarks and our probing benchmarks.}
    \label{fig-ft-results}
    \vspace{-.15in}
\end{figure}

Although the main purpose of \method is to evaluate \llms, its generated samples can also be used as augmented data for fine-tuning.
In this section, we conducted a pilot study using the OpenAI API to fine-tune \gptthree on the data generated by \method.
Specifically, we used samples from the training split of \mmlu and \arc, which are fed to \method for data generation.
We then evaluated the fine-tuned models on the original test split and our probing benchmarks. The fine-tuning data includes two parts, the first part is the probing questions generated by $5$ principles separately, and the second part is the original training set.

The results in \cref{fig-ft-results} show that the data generated by \method can improve the performance of \llms, with an average of $2\%$ improvements on both \mmlu and \arc.
The fine-tuning results demonstrated that \method is not only an evaluation protocol, but a general data augmentation approach to improve the performance of \llms, creating a huge advantage for training stronger \llms in the future.
The improved results also demonstrate the correctness of the generated data, indicating that our \method is effective.

\section{Conclusion and Discussion}
This paper introduced \method, a dynamic evaluation protocol to address data contamination and provide an in-depth analysis of the three key cognitive abilities of \llms inspired by psychometric theory. Our experimental findings revealed several notable insights. Crucially, \method-generated samples can not only function as evaluation tools, but also improve \llms training as a data augmentation method.
We believe that the psychometric-inspired adoption of \llms as agents represents a promising direction.

Our work has several limitations.
(1) Tasks and datasets: Our focus was limited to four datasets, encompassing a specific range of topics. Incorporating a broader spectrum of datasets and tasks could yield more comprehensive insights into \llms capabilities.
(2) The validity of probing benchmarks: While \method employs a judge agent to assess the consistency and accuracy of probing benchmarks, we observed discrepancies in some questions, deviating from their original intent. This highlights the potential to further enhance \method's robustness and effectiveness.

\section*{Impact Statement}
Evaluating the general abilities of \llms is essential to ensure responsible AI for the society.
This work proposed a new evaluation protocol of \llms to ensure that their true capabilities can be measured, which will help foster a better understanding of the models.
We carefully controlled the generative models (agents) in the paper to ensure that they will not generate harmful content.

\bibliography{ref}
\bibliographystyle{icml2024}

\newpage
\appendix
\onecolumn

\section{Datasets}
\label{append-dataset}
The \mmlu dataset contains $13,985$ test samples across $57$ tasks, encompassing diverse areas such as humanities and social sciences, offering a comprehensive assessment of language understanding capabilities. The \arc dataset collected $1,172$ grade-school level science questions, presenting a unique blend of natural language understanding and scientific reasoning. In the GSM8K dataset, the focus is on mathematical problem-solving, featuring $1,319$ problems that require a combination of numerical understanding and logical reasoning. For the Formal Fallacies, Object Counting, and Temporal Sequences tasks in \bbh dataset, each contains $250$ test samples. These subsets were chosen for their relevance and representativeness, as they challenge \llms to understand nuanced logical fallacies, accurately count objects in complex settings, and understand sequences of events over time.

\section{Evaluation Prompts}
\label{append-eval-prompts}

In the following, we show the evaluation prompts while adopting different datasets.

\textbf{MMLU} 

\prompt{Here is a question about \{task\}:

\{question\}

\{choices\}

Choose the correct answer and explain why. Please include your answer into <<<>>>. For example, if you choose A, please write <<<A>>>.}

\textbf{GSM8K}

\prompt{Here is a math problem:\{question\}

Please solve this math problem and include your answer into <<<>>>. For example, if your answer is 1, please write <<<1>>>.}

\textbf{ARC-C}

\prompt{
Here is a multiple-choice science problem:

\#\#\# Question:

\{question\}

\#\#\# Choices:

\{choices\}

Please solve this problem and include your answer into <<<>>>. For example, if your choose A, please write <<<A>>>.}

\textbf{BBH (formal fallaices)}

\prompt{
Here is a question about formal fallacies (given a context involving a set of statements, determine whether an argument can be logically deduced from the provided context):

\#\#\# Question:

\{question\}

\#\#\# Choices:

\{choices\}

Please answer this question and include your answer into <<<>>>. For example, if your answer is valid, please write <<<valid>>>.
}

\textbf{BBH (object counting)}

\prompt{
Here is a question about object counting (given a collection of possessions that a person has along with their quantities, determine the number of a certain object/item class.):

\{question\}

Please answer this question and include your answer into <<<>>>. For example, if your answer is 1, please write <<<1>>>.
}

\textbf{BBH (temporal sequences)}

\prompt{

Here is a question about temporal sequences (given a series of events and activities a person has completed in the course of a day, determine what time, during the day, they might have been free to perform another activity.):

\#\#\# Question:

\{question\}

\#\#\# Choices:

\{choices\}

Please answer this question and include your answer into <<<>>>. For example, if your answer is (A), please write <<<(A)>>>.
}

\section{Detailed Results}
\label{sec-append-detailed-results}

\subsection{Standard Deviation of Main Results}
\label{sec-append-std}

The dynamic evaluation protocol introduces randomness into the evaluation results. Therefore, we run all experiments three times to get the average results and the standard error.
As shown in \cref{table: std}, the standard deviations for all models in all data sets are small, thus ensuring the fairness of our evaluation.

\begin{table}[htbp]
\centering
\caption{The standard deviation of co-efficient of the main results.}
\label{table: std}
\resizebox{0.45\textwidth}{!}{
\begin{tabular}{ccccc}
\toprule
  Model         & MMLU       & GSM8K      & ARC-C      & BBH (partial)        \\
\midrule
\gptfour  & 0.25 & 0.89  & 0.46  & 1.95 \\
\gptthree & 0.18 & 1.15  & 0.58  & 1.49 \\
\gemini   & 0.13 & 1.69  & 0.58  & 2.89 \\
\yi       & 0.05 & 1.63  & 0.58  & 2.51 \\
\mixtral  & 0.62 & 0.16  & 0.73  & 2.37 \\
\llama    & 0.93 & 1.52  & 1.12  & 1.42 \\

\bottomrule
\end{tabular}
}
\end{table}

\subsection{Results of Different Modular Principles}
\label{sec-apped-ablation-principles}

We show the results on different principles in \cref{table: ablation principle 1} and \cref{table: ablation principle 2}.
Note that we only adopted partial samples from BBH.

\begin{table*}[htbp]
\parbox{0.6\linewidth}{
\caption{Results of different principles on MMLU and ARC-Challenge datasets.}
\label{table: ablation principle 1}
\resizebox{0.6\textwidth}{!}{
\begin{tabular}{cccccccc}
\toprule

Dataset    & Model      & Baseline & $p_1$     & $p_2$     & $p_3$     & $p_4$     & $p_5$     \\ \midrule
\multirow{3}{*}{MMLU}          & GPT-4      & 84.40    & 78.43 & 81.48 & 80.27 & 81.42 & 83.73 \\
                               & GPT-3.5    & 68.12    & 64.71 & 67.39 & 64.31 & 64.61 & 67.09 \\
                               & Gemini-Pro & 67.04    & 64.28 & 66.27 & 63.89 & 62.77 & 64.71 \\ \midrule
\multirow{3}{*}{ARC-Challenge} & GPT-4      & 96.16    & 94.11 & 94.28 & 93.69 & 93.69 & 96.50 \\
                               & GPT-3.5    & 85.41    & 84.39 & 84.64 & 83.02 & 82.25 & 85.67 \\
                               & Gemini-Pro & 86.18    & 84.13 & 84.81 & 83.36 & 81.83 & 85.24 \\

\bottomrule
\end{tabular}
}
}
\hfill
\parbox{.4\linewidth}{
\caption{Results on GSM8K and BBH (partial) datasets.}
\label{table: ablation principle 2}
\resizebox{0.375\textwidth}{!}{
\begin{tabular}{ccccc}
\toprule
Dataset & Model      & Baseline & $p_1$     & $p_2$     \\ \midrule
\multirow{3}{*}{GSM8K}      & GPT-4      & 95.22    & 90.83 & 91.66 \\
                            & GPT-3.5    & 77.71    & 74.98 & 74.07 \\
                            & Gemini-Pro & 22.97    & 23.58 & 22.52 \\ \midrule
\multirow{3}{*}{BBH (partial)}        & GPT-4      & 88.53    & 89.60 & 89.47 \\
                            & GPT-3.5    & 54.67    & 52.27 & 48.40 \\
                            & Gemini-Pro & 65.47    & 66.93 & 61.73
\\

\bottomrule
\end{tabular}
}
}
\end{table*}

\subsection{Results of Relationship of the Basic Abilities}
\label{sec-append-relation}

\cref{tb: complexity} shows the results on different abilities.

\begin{table}[t!]
\centering
\caption{Results of different \llms on \method based on ARC-C and MMLU datasets.}
\label{tb: complexity}
\resizebox{\textwidth}{!}{
\begin{tabular}{c|ccc|ccc}
\toprule
Dataset & \multicolumn{3}{c|}{ARC-C}                                      & \multicolumn{3}{c}{MMLU}                                       \\ 
Model &
  \multicolumn{1}{c}{Language understanding} &
  \multicolumn{1}{c}{Problem solving} &
  Domain knowledge &
  \multicolumn{1}{c}{Language understanding} &
  \multicolumn{1}{c}{Problem solving} &
  Domain knowledge \\ \midrule
\gptfour   & \multicolumn{1}{c}{90.27} & \multicolumn{1}{c}{94.28} & 93.69 & \multicolumn{1}{c}{75.18} & \multicolumn{1}{c}{81.48} & 81.42 \\ 
\gptthree   & \multicolumn{1}{c}{79.18} & \multicolumn{1}{c}{84.64} & 82.25 & \multicolumn{1}{c}{61.02} & \multicolumn{1}{c}{67.39} & 64.61 \\ 
\gemini  & \multicolumn{1}{c}{80.46} & \multicolumn{1}{c}{84.81} & 81.83 & \multicolumn{1}{c}{59.53} & \multicolumn{1}{c}{66.27} & 62.77 \\ 
\yi      & \multicolumn{1}{c}{79.44} & \multicolumn{1}{c}{85.67} & 83.19 & \multicolumn{1}{c}{60.01} & \multicolumn{1}{c}{66.10} & 64.50 \\ 
\mixtral & \multicolumn{1}{c}{78.16} & \multicolumn{1}{c}{82.17} & 78.58 & \multicolumn{1}{c}{61.18} & \multicolumn{1}{c}{66.01} & 61.64 \\ 
\llama   & \multicolumn{1}{c}{70.14} & \multicolumn{1}{c}{75.00} & 68.94 & \multicolumn{1}{c}{54.54} & \multicolumn{1}{c}{57.60} & 54.23 \\ \bottomrule
\end{tabular}
}
\end{table}

\subsection{Top topics of MMLU}
\label{sec-append-topics}

\cref{fig-error-top-20} shows the top 20 MMLU topics where \gptfour made the most errors. It can be observed that \gptfour made more mistakes in ``profession law'', ``moral scenarios'', and ``security studies'', potentially due to insufficient training data and ambiguious groundtruth in these domains.
For example, questions from ``moral scenarios'' are often difficult to answer.
This trend underscores potential limitations in \gptfour's current understanding or processing capabilities with respect to the ethics and psychology domains.

\begin{figure}[htbp]
    \centering
    \includegraphics[width=0.55\textwidth]{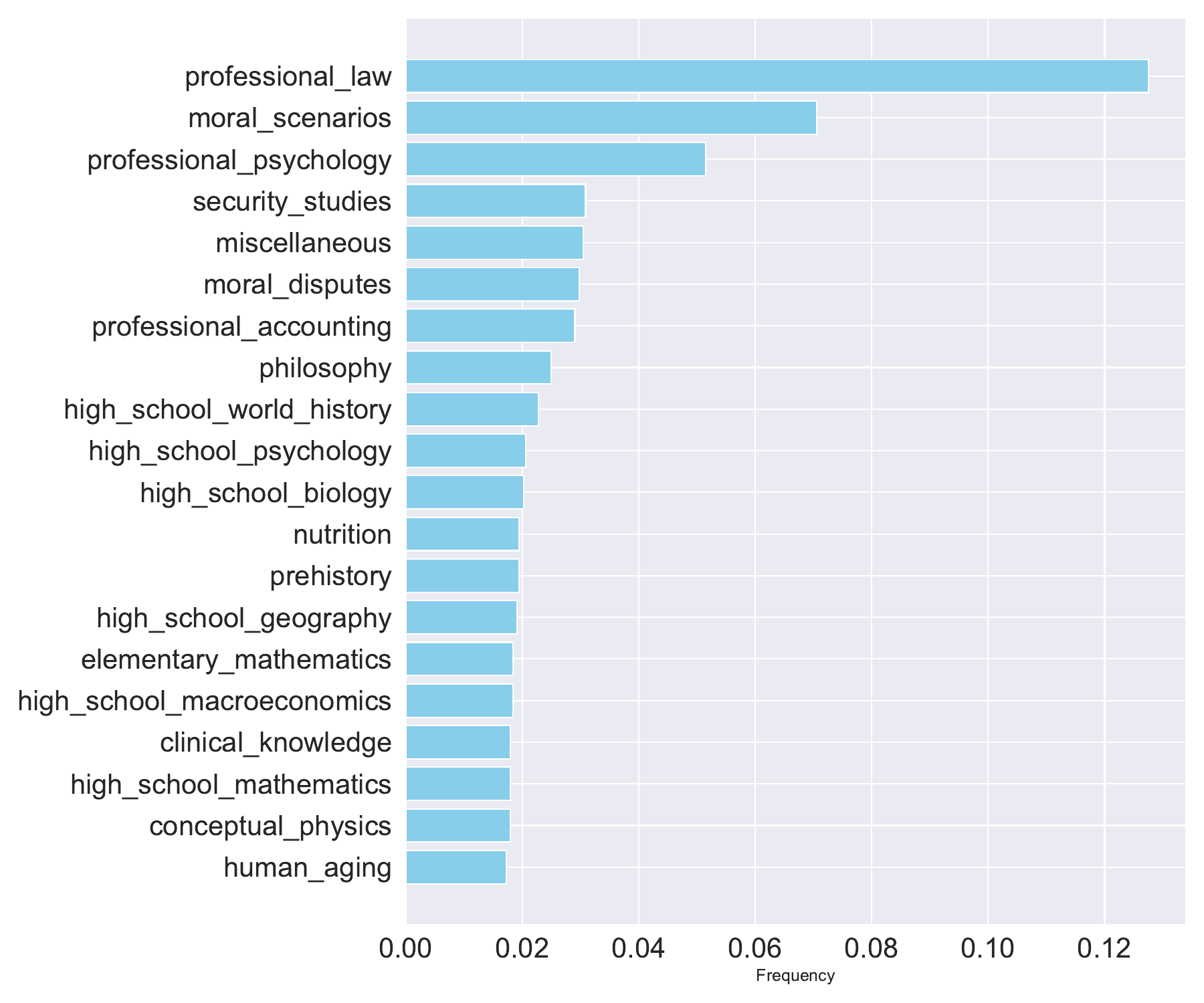}
    \caption{The bar chart of top 20 topics and their corresponding frequencies of \gptfour on \mmlu dataset.}
    \label{fig-error-top-20}
\end{figure}

\subsection{Examples of wrong/ambiguous evaluation samples}
\label{sec-append-ambiguous-examples}

In this section, we presented several wrong/ambiguous evaluation samples in \gsm dataset.
\begin{itemize}
    \item \textbf{Question:} \prompt{Lee used to be able to run the 400-meter hurdles two seconds faster than Gerald would run the 400-meter hurdles.  But Gerald changed his diet, which improved his speed by 10\%.  If Lee runs the 400-meter hurdles in 38 seconds, how fast can Gerald, with his improved diet, run the 400-meter hurdles, in seconds?}
    \item \textbf{Answer:} $36$
    \item \textbf{Analysis:} The correct answer is $36.3636$.

    \item \textbf{Question:} \prompt{Mandy owes Benedict \$100. They agreed to have monthly interest of 2\%. If Mandy was able to pay it after 3 months, how much should she give to Benedict?}
    \item \textbf{Answer:} $106$
    \item \textbf{Analysis:} The financial arrangement between Mandy and Benedict involves a principal loan of \$100 with an agreed monthly interest rate of 2\%. The ambiguity in the original question arises from the lack of specificity regarding the interest calculation method: simple or compound. The provided answer (\$106) initially suggests a simple interest calculation. However, considering the possibility of compound interest sheds light on a different approach to determining the final amount owed.
    
\end{itemize}

\section{Examples generated by \method}
\label{sec-append-examples}

Finally, we show some examples generated by \method on the basis of different benchmarks.

\paragraph{\mmlu:}
\begin{itemize}
    \item \textbf{Original Question:} \prompt{This question refers to the following information.
    
    Read the the following quotation to answer questions.
    
    The various modes of worship which prevailed in the Roman world were all considered by the people as equally true; by the philosopher as equally false; and by the magistrate as equally useful.
    
    Edward Gibbon, The Decline and Fall of the Roman Empire, 1776–178
    
    Gibbon's interpretation of the state of religious worship in ancient Rome could be summarized as
    
    A: In ancient Rome, religious worship was decentralized and tended to vary with one's social position.
    
    B: In ancient Rome, religious worship was the source of much social tension and turmoil.
    
    C: In ancient Rome, religious worship was homogeneous and highly centralized.
    
    D: In ancient Rome, religious worship was revolutionized by the introduction of Christianity.
    
    Answer: A}
    \item \textbf{Probing Question:} \prompt{In his seminal work ``The Decline and Fall of the Roman Empire,'' Edward Gibbon explores the vast expanse of Roman history, from its zenith to its eventual collapse. Amid his detailed examination, Gibbon provides insights into the cultural and religious diversity that characterized the Roman Empire. How does he describe the differing perspectives of the general public, philosophers, and government officials regarding this religious diversity, particularly in terms of their acceptance and the impact on Roman society?
    
    A: Religious worship in ancient Rome was uniform and controlled by a central authority.
    
    B: The multiple forms of religious worship in ancient Rome often led to social conflicts and disturbances.
    
    C: The Roman state endorsed all forms of worship equally in an attempt to appease the gods and ensure the empire's prosperity.
    
    D: The arrival of Christianity in ancient Rome was a transformative force that completely changed the nature of religious worship.
    
    E: Religious practices in ancient Rome were not centralized, and they varied according to the social status of an individual.
    
    Answer: E}
\end{itemize}

\paragraph{\gsm:}
\begin{itemize}
    \item \textbf{Original Question:} \prompt{Janet’s ducks lay 16 eggs per day. She eats three for breakfast every morning and bakes muffins for her friends every day with four. She sells the remainder at the farmers' market daily for \$2 per fresh duck egg. How much in dollars does she make every day at the farmers' market?
    
    Answer: 18}
    \item \textbf{Probing Question:} \prompt{Janet has a small farm where she raises a variety of animals, but her ducks are the most productive when it comes to laying eggs. Each day, without fail, her flock of ducks provides her with 16 fresh eggs. Janet has a particular routine where she enjoys a hearty breakfast that always includes three scrambled eggs. After breakfast, she dedicates some time to baking, preparing four delicious muffins that she shares with her friends. These muffins are special because they each require one egg. After using the eggs for her breakfast and baking, Janet packages the remaining eggs to be sold at the bustling local farmers' market. Her eggs are quite popular, and she sells them for \$2 each. Given her daily routine, how much money does Janet make from selling her eggs at the farmers' market each day?    
    
    Answer: 18}
\end{itemize}

\paragraph{\arc:}
\begin{itemize}
    \item \textbf{Original Question:} \prompt{An astronomer observes that a planet rotates faster after a meteorite impact. Which is the most likely effect of this increase in rotation?
    
    A: Planetary density will decrease.
    
    B: Planetary years will become longer.
    
    C: Planetary days will become shorter.
    
    D: Planetary gravity will become stronger.

    Answer: C
}
    \item \textbf{Probing Question:} \prompt{In the vast expanse of the solar system, where celestial bodies are constantly in motion, a planet's day-to-day existence can be altered by events such as collisions with other objects. Imagine a scenario where astronomers witness a planet whose day has been significantly shortened due to the impact of a meteorite. This incident has resulted in the planet having a quicker rotational period around its axis. Given this situation, what is the probable consequence of this accelerated spin on the planet's environment or physical state?
    
    A: The time it takes for the planet to orbit the sun will increase.
    
    B: The duration of a single rotation of the planet on its axis will be less.
    
    C: The planet's atmosphere will become significantly thicker due to increased centrifugal force.
    
    D: The force with which the planet pulls objects towards itself will intensify.
    
    E: The mass per unit volume of the planet will be reduced.

    Answer: B
    }
\end{itemize}

\paragraph{\bbh (formal fallacies):}
\begin{itemize}
    \item \textbf{Original Question:} \prompt{Here comes a perfectly valid argument: First of all, whoever is a schoolmate of Sondra is not a stepsister of Pricilla. In consequence, whoever is not a stepsister of Pricilla is a schoolmate of Sondra.
    
    Is the argument, given the explicitly stated premises, deductively valid or invalid?
    
    Options:
    
    - valid 
    
    - invalid
    
    Answer: invalid
    }
    \item \textbf{Probing Question:} \prompt{At Ridgemont High, a peculiar rule is established by the student council: anyone who is a classmate of Sondra cannot concurrently be a half-sibling of Pricilla. This rule came about after a complex dispute over club memberships and family connections within the school. Now, consider a debate that erupted in the philosophy club during a discussion on logical reasoning. One of the members presented what seemed to be a sound argument based on the student council's rule: if it is true that no classmate of Sondra can be a half-sibling of Pricilla, then it logically follows that anyone who is not a half-sibling of Pricilla must be a classmate of Sondra. The club is now pondering whether this argument is logically coherent and deductively sound, based on the premises provided by the student council's peculiar rule.
    
    Options:
    
    - valid 
    
    - invalid

    Answer: invalid
    }
\end{itemize}

\paragraph{\bbh (object counting):}
\begin{itemize}
    \item \textbf{Original Question:} \prompt{I have a flute, a piano, a trombone, four stoves, a violin, an accordion, a clarinet, a drum, two lamps, and a trumpet. How many musical instruments do I have?
    
    Answer: 8}
    \item \textbf{Probing Question:} \prompt{As a passionate collector and music enthusiast, I've dedicated a significant portion of my living space to housing various items that reflect my interests and hobbies. Over the years, I've amassed a collection that includes both musical instruments and household items. Among my cherished possessions are one flute, one piano, one trombone, one violin, one accordion, one clarinet, one drum, and one trumpet. In addition to these, my practical side has led me to acquire four stoves and two lamps to meet my daily needs. Given this eclectic mix of items, can you carefully count and tell me how many items from my collection are musical instruments?
    
    Answer: 8}
\end{itemize}

\paragraph{\bbh (temporal sequences):}
\begin{itemize}
    \item \textbf{Original Question:} \prompt{
    Today, Susan went to the coffee shop. Between what times could they have gone?
    
    We know that:
    
    Susan woke up at 7am.
    
    Linda saw Susan driving to the water park from 7am to 11am.
    
    John saw Susan buying clothes at the mall from 11am to 12pm.
    
    Jessica saw Susan taking photos near the Eiffel Tower from 12pm to 1pm.
    
    Steven saw Susan buying lunch at the deli from 1pm to 2pm.
    
    Thomas saw Susan reading at the library from 2pm to 6pm.
    
    The coffee shop was closed after 9pm.
    
    Between what times could Susan have gone to the coffee shop?
        
    Options:
    
    (A) 6pm to 9pm
    
    (B) 7am to 11am
    
    (C) 1pm to 2pm
    
    (D) 2pm to 6pm

    Answer: (A)
    }
    \item \textbf{Probing Question:} \prompt{On which occasion during her busy schedule could Susan have potentially squeezed in a visit to the local coffee shop? Susan's day kicked off at the crack of dawn, at 7am. Between the early hours of 7am and the late morning at 11am, Linda witnessed Susan making her way to the refreshing water park, where she was set to enjoy the slides and pools. During the late morning hour, from 11am to noon, John caught a glimpse of Susan amidst the bustling shoppers at the mall, where she was selecting some fashionable clothing items. As the clock struck noon and the day progressed to 1pm, Jessica was with Susan, snapping pictures against the backdrop of the iconic Eiffel Tower at a well-visited tourist spot. Following her tourist escapades, from 1pm to 2pm, Steven saw Susan at the cozy deli downtown, where she was deciding on her midday meal from a variety of savory options. Later in the afternoon, from 2pm to 6pm, Thomas noticed Susan deeply engrossed in literature at the quiet library, a place where she often finds solace in the pages of her favorite books. It's also important to note that the coffee shop in question shuts down its espresso machines and locks its doors to customers promptly at 9pm. Given Susan's known whereabouts throughout the day, deduce the time interval where she could have enjoyed a coffee shop visit.
    
    Options:
    
    (A) 6pm to 9pm
    
    (B) 7am to 11am
    
    (C) 1pm to 2pm
    
    (D) 2pm to 6pm

    Answer: (A)
    }
\end{itemize}

\section{Human verification results}
\begin{table}[!ht]
\caption{Results of human verification on MMLU and ARC-Challenge datasets.}
\label{tb: appendix-human}
    \centering
    \begin{tabular}{ccccc}
    \toprule
        Equivalence / Correctness & Language Understanding & Problem Solving & Domain Knowledge & Avg \\ \midrule
        Group 1 & 0.88/0.95 & 0.91/0.93 & 1.00/1.00 & 0.93/0.96 \\ \midrule
        Group 2 & 0.92/0.99 & 0.94/0.98 & 1.00/0.97 & 0.95/0.98 \\ \midrule
        Group 3 & 0.91/0.95 & 0.92/0.99 & 1.00/1.00 & 0.94/0.98 \\ \midrule
        Avg & 0.90/0.96 & 0.92/0.97 & 1.00/0.99 & 0.94/0.97 \\ \bottomrule
    \end{tabular}
\end{table}

\end{document}